# Law of Connectivity in Machine Learning

Jitesh Dundas
Scientist, Edencore Technologies (www.edencore.net)
Row House – 6, Opp Ambo Vihar, Tirupati Nagar-II, Off Unitech Road, Virar(w)
Thane-401303, Maharashtra, India
jbdundas@gmail.com

*Abstract*— **We present in this paper our law that there is always a connection present between two entities, with a self-connection being present at least in each node. An entity is an object, physical or imaginary, that is connected by a path (or connection) and which is important for achieving the desired result of the scenario. In machine learning, we state that for any scenario, a subject entity is always, directly or indirectly, connected and affected by single or multiple independent / dependent entities, and their impact on the subject entity is dependent on various factors falling into the categories such as the existence of entity, the inner state of the entity, the external state of the entity and the state of communication of the entity.**

I. INTRODUCTION

In this paper, we present our work on the relationship between the ability of the entities to create connections and the different reactions that may generate from them. We present the above law in the following manner:-

$$\text{Distance }(E_n) = f(X) = \sum_{n=0}^{\infty} \{E_{(n-1)} \longleftrightarrow E_{(n)}\} > 0 \qquad (2)$$

Where $E_{(n)}, E_{(n-1)} > 0$ and E is the entity (with index 'i') attributes (mentioned above) which are involved in the interaction. We postulate that the required attributes may be added further, depending upon the requirements of the learning algorithms, besides the one shown above. Thus, we measure them as $E_i$ in the above equation. The above equation can serve as a neural relaxation model and thus we further derive the equation to be:-

$$\text{Distance }(E_n) = f(X) = \sum_{n=0}^{\infty} (E_{(n)} / E_{(n-1)}) * (E_{(n-1)} / E_{(n-2)}) \ldots \ldots \ldots * (E_{(1)} / E_{(0)}) \qquad (3)$$

Thus, $E_n > 0$ and $1 > E_i > 0$. A value of 0 defines that the interaction is not possible due to the entity's inability to pursue the interaction. Again, the value can never been 1 as there is always some disturbance or obstacle that will hinder the interaction. Again, the measurement is probabilistic and $E_n > 0$ and $E_n < 1$. A value of 0 defines that the interaction is not possible due to the entity's inability to pursue the interaction. Again, the value can never been 1 as there is always some disturbance or obstacle that will hinder the interaction. Again, the measurement is probabilistic and $E_n > 0$ and $E_n < 1$, where $E_n$ is the intended initiator and $E_0$ is the intended recipient of the interaction. The entities considered in the interaction depend on the path that is chosen for learning and considering this interaction. The initiator may a
take a mirror and look at the image of the recipient, or talk to that person over the phone or even go personally to meet the recipient. Every path chosen, depending on the learning algorithm, will profoundly impact the selection of entities and thus decide the quality and time of the interaction. The number of entities varies in any learning scenario and thus plays an important role in the quality of interaction.

This law will make the foundation of neural networks, decision tree based techniques and other learning algorithms. This law will allow the implementation of long as well as facilitate deeper investigations of networks in learning models. Since any model is prone to connection with some other entity, it has to decide whether it wants to connect or not. This decision decides whether the interaction will happen or not. The absence of interaction between two entities means that the path to connect the entities is being ignored or is difficult to follow. However, we postulate that the connection path always exists but the ability of the entities to become aware and use the connection is what is missing. The latter decides the scenario and the scope of the interaction too. One of the most important problems in machine learning is teaching the computer to observe [1]. There are certain high levels of functions that humans do better than computers such as creativity, observation and imagination. Of all these functions, observation is of vital interest for machine learning as this ability is at the root of all the high levels of human functions. Observation is defined as the ability to understand and interpret the inner capabilities and unravel the complex functions of the entities under observation of the computer's interaction environment. For anything to be learned by the computer, it must be able to establish a connection with it. This may happen in a series of steps or connections that will allow us to reach to our goal as shown in Figure-1. Which one will the computer takes depends on the ability to create the context and its interests and the desired end result. Any path or algorithm used will need connectivity, besides other factors, to come up with a path from Entity A-> Entity B will need connectivity and fulfill the needed parameters, to be able to learn and conceive the desired the results. The human brain establishes patterns and forms all the operations based on its understanding of the entities in its environment. In order to do that successfully, the computer needs to establish proper channels or paths towards the desired entity. Learning is a phased series of steps, which will take the computer towards a higher level of knowledge and existence. This activity allows the intended recipient to become well-versed with the environment. For e.g. if a person wants to look at another person in the adjacent building, he must be able to connect with the latter in some manner, in reality or in his imagination. In short, he tries to connect with the person in some way. In order to fully achieve this, he has to satisfy certain parameters that will get him this:-

1) **Existence of entity: -** The entity that the person desires to achieve is present in his context or scenario under observation. The other person he desires to connect too must exist or reach him via vision, audio or other even imagination. The entity surely exists in his scenario in this example. If not, then this learning is not possible using any of the known learning algorithms or in any other methods.

2) **Inner state of the entity: -** The state of the entity plays a huge role in deciding whether the connection, that will ensure the interaction, will exist or not. If the other is not speaking or not even visible to this person, then the desired interaction will not happen. Clearly, the connectivity is at the centre of the desired learning or even interaction with the other entities. This connectivity will depend on the inner state of the entities. There is an imminent need to have both the entities internally ready to have the interaction. The desire and the intent of the entities should be positive and as per the formulae $0 > E_i > 1$.

3) **External state of the entity: -** The external state refers to the entity's ability to pursue in its physical form, the interaction. The entity may be facing obstacles form other entities or from the environment. It might be also getting help or using other entities to pursue its connectivity with the intended recipient. This external state also is important in concluding whether the entity is ready to communicate with the other entities. Is it having the required clearance and path to achieve its goals? Is there the required knowledge or presence of entities that will give it the same, to be able to reach its goals? The external state also considers the kind of synchronization it has with the environment. Is the entity in line or in proper condition to be able to pursue this interaction? If the state is positive, only then will the interaction occur. Imagination is also the external state condition in which the measurement of this attribute goes to near 0. This interaction will this fail and have very less accuracy as it does not ensure the presence or the desired results. The path selected in the interaction must be overlapping the path of the interaction in the real scenario. For e.g. A person who imagines that the sun is rising in the east must be in line with the fact that the sun rises in the east. If the sun rises in the east and the entity (machine) expects (in its connection via its learning algorithm) that the sun rises in the west, then the latter's state is not in sync with the reality and will thus fail the scenario tests. The machine is thus entitled to conditions such as confusion, ignorance and degradation of knowledge and incorrect result occurrence. **The presence or absence of obstacles will decide the fate of the interaction using the path between the entities.** The entities will always have atleast one path between them, but will be able to pursue this only if the four entity attributes (or more) are present and fulfilled properly. Imagination and dreams are hypothesis that the brain creates to present the state of the entity, in relation to the environment. Imagination is the result of the obtained interaction or the feedback the entity (human being in this case) obtained from its previous interaction. However, dreams are nothing but the desires or expectations that the entity wishes to obtain using the learning algorithms.

4) **State of communication of the entity**:- The interaction must be free from obstacles till the entity has fulfilled its desired results. The result will be dependent on the ability of the entities to carry out this interaction. There are switches in the entities, which when in the ready state, ensure that the communication occurs. These are nothing but intentions or motives that are present in the entities. If there is ignorance or lack of interest or any such state present, then the communication will not occur.

As per the law in this paper, there is always a connection between two entities. Thus, the hidden and unknown entities can be connected using a silent connection (it is a connection that may not be present in reality but it can give a possible connection between the entities. For e.g. we hypothesize that an unknown person may have caused defaults whereas in reality there is no connection to prove that. In short, there was an imaginary connection but there was no overlap to the reality that was discovered later. We call such connections as silent connections)

## II. BACKGROUND

There are several algorithms that exist in today's machine learning literature. Till date, no such fact or equation exists that ensures the base for creating or writing learning algorithms. The learning algorithms thus remain at a threat of being away from their intended purpose and inapplicable in certain states. There is no single algorithm for different purposes or conditions. However, the certain attributes and the method of pursuing the learning algorithms will always remain the same. We have studied all the algorithms mentioned in the Methods section and have found that they clearly have dependencies on the law of connectivity. They use the Connectionists approach [5, 6], but this approach does not mention anything on the parameters and the specific details that affect the interaction using the learning algorithms. The law of connectivity clearly fills the gap in this direction, establishing the base that is needed for the algorithms mentioned and studied here [8]. Also, there is no mention of the fact that there is always a connection present between two or more entities, in any form or state as needed. It is the obstacles that prevent the interaction or the learning process from obtaining the desired results of learning. Decision Tree based algorithms require the presence of tree based approaches towards achieving the desires learning interaction. However, the connectivity is at the centre of these algorithms as tree based approaches depend on the connection of the branches of its scenario. Supervised learning techniques require the path to be laid before hand (for e.g. learning by imitation) so that the intended initiator can learn and achieve the desired recipient. Unsupervised techniques such as Adaptive Neural networks [2] create connections between entities in order to move towards their solution. They have been successful in solving several problems in machine learning [3]. However, they fail to obtain the required accuracy as in humans as they do not fully implement the law of connectivity. The problem of loan calculations and risk assessment requires that the states of quality of communication (or the presence of clear logic) be present. Also, the number of entities that shape the desired interaction (the presence of all the facts which are entities in this case) is essential to obtain the correct interaction. In complex problems, the law of connectivity is needed to be fully implemented to obtain the desired results. Bayesian learning [7, 11] does not consider the modes connectivity approach with the parameters and their effects. Some of the algorithms consider the assignment of weights but do not specify the condition and measurement of the same. Prior prediction of the states and the inability of the theorem to be practically applied are major hindrances in using these Bayes theorem based techniques. Identification of states and the acceptance of the dynamic modes of the entities as well as the environment are also missing in them. Similar states were found to be true for other similar algorithms present in this category too.

Hidden Markov Models [9] or HMM were found to consider hidden states of the entities in the scenario and measure their impact by assigning them weights. However, this model fails to consider the case of unknown entities i.e. entities about which we have no information except for the effect that they have on the communication in the scenario. This paper explains this void with the help of silent connections and shows how connections always exist between two entities. Boltzmann machine [10] based algorithms fail to solve complex scenario based computations and are known to be less practical. Machine Learning algorithms have to take into account actions such as inference [12], imagination and creativity, some of which require prior knowledge. Cased based learning [13] requires the knowledge of past experience to solve the current problems in learning, which is less productive for the case of unknown entities. Hidden entities and distractions [14] are very commonly found in any normal learning scenario. Most of the supervised learning algorithms revolve around solving these issues in one way or the other [15]. Clustering based algorithm [16] such as QT algorithm need to cluster the entities into subsets, but they do not talk of how unknown entities and external entities could affect the learning. Expectation-Maximization algorithms such as maximum likelihood and k-means algorithm [12, 17] handle unknown entities to some extent (by iterating over the existing data and building on the new generated results to complete the missing information and obtain the required information) but fail to consider the switches and the multiple connections that may be present in between two entities. They also fail to provide coverage for incorrect or false connections, measuring the correctness and the resulting impact on the scenario. The silent connections need to be considered but is not the case in the former. Temporal Difference Learning [18] and Self-Organizing Map [19] also follow the relative and iterative technique of learning about unknown entities. They are better in establishing connections but fail to handle coverage and the quality aspects to allow connections in the scenario. Associative rule based techniques such as Apriori [20] algorithm requires the creation of associative rules to create data about the scenario, which could be less useful in the case of unknown entities, especially since their connections themselves are silent in most cases. In an excellent paper by Tishby et al [21], the accuracy and complexity are compromised to find out the probabilistically best information for the scenario under investigation. However, there is a lack of information on coverage and quality of interaction. The focus is on the distance and clusters that are involved rather than the quality of the connections and the coverage involved. The tradeoffs ensure that the required learning levels may be missed out in fuzzy cases, especially in which unreliable information is present. IBSEAD [22] is an unsupervised learning algorithm that handles unknown entities better then neural networks due to the presence of better coverage and condition based interaction. This algorithm is very useful in handling complex situations and novel scenarios where no information about the entities (external or internal) is present except for their effect on the interaction. The work by Amarel on the representation of entities as a state space search [23] was pioneering with effects on better techniques for path representation. However, no work has been done on equations for neural relaxation models. The equations in this paper serve to be the first step in this direction. Newall, Shaw and Simon [24] at CMU proposed the "no single algorithm" on p. 5 which showed the failure of the general problem solver (GPS) algorithm. The reason was the combinatorial expansion of states in between the source and destination. Practical Reasoning [25] requires the ignorance of larger scenario and consideration of a few entities only. The mind –body relationship is to be understood deeply in order to be able to provide a finer analysis of the ways in which machines could be made better in higher levels of human abilities [26]. The work presented in this paper is in line with the work done by Kendel [27] in understanding and creating a base for machine learning and mind studies. The study of the mind is much more than the just the biological processes and thus biological studies will not be able to fully appreciate and help in finding deeper insights into the workings of the human mind (which is essential for giving machines the intelligence of the human brain levels). However, this mode of practical reasoning only allows for loss of information and less coverage will lead to higher inaccuracies and unnecessary fuzziness. [28] Hegel's triad was also very important in helping carve the modes of analysis in contemporary analysis. However, this triad method was found to be erraneous later and discarded by Hegel himself. We also tend to ignore the presence of network maps in any scenario due to convenience and common human behaviour, something that is due to convenience and lack of interest for accuracy. This law can lays down the concepts of coverage, interaction and silent connections which are instrumental in assuring the success of learning algorithms and techniques. Most of the drawbacks in the existing algorithms are because of their inability to satisfy these attributes. In any of the interactions in the learning algorithms, the satisfactory implementation of the law of connectivity is essential to have the successful impact and result of the learning in machines. If an entity initiator is unable to do the interaction in reality, then even such a case, the entity will have to, directly or indirectly, using imagination or using fallacy, create a scenario in which the interaction exists. The accuracy and the feedback from the resulting interaction will decide if the observed imaginative interaction did produce any accurate or real-time results. The law of connectivity will thus decide whether the algorithm was able to achieve the desired results or not. If the parameters are fully appreciated in the learning of the entity, then the desired interaction will be implemented to the best possible extent.

### III. FORMULAE

In the earlier equation – (2), we stated that:-
$$\text{Distance } (E_n) = f(X) = {}_{n=0}\sum\nolimits^{\infty} \{E_{(n-1)} <--->E_{(n)}\} > 0 \qquad (4)$$

Where Distance = Connection (or Conn) i.e. C1, C2 between the two entities. We have postulated here that one connection has at least two entities and each of the two entities (or more) can have more than one connection between them. This is to accommodate different behaviors and effects between the same entities at different or at the same time. For e.g., Entity A is interacting with Entity B

Connectivity = $C_1 (+-) C_2 (+-) \ldots\ldots (+-) C_N$     (5)

Here, each of the Connections C1, C2, etc may give negative values if their impact is hindering the efficiency of the system learning. If the connection improves the latter, then a positive value is given. Each connection Ci is given by:-

$C_i = F(E_i <==> E_j) = \sum_{t=0}^{\infty} \sum_{i=0}^{n} \sum_{j=0}^{m} \sum_{k=0}^{p} \sum_{l=0}^{q} ((Imp) * E_{(i,k)} -- (Imp) * E_{(j,l)})$     (6)

Where,
Conn = Connection between the entities.
N = number of connections.
t = time of the scenario in the dynamic system
i = index of the source entity
j = index of the destination entity
k = index of the connection from source to destination. Two entities may have more than one connection between them.
l = index of the connection from destination to source. Two entities may have more than one connection between them.
Imp = Impact factor averaged out with all the attributes considered and measured. Imp is the **impact factor** based on all the attributes considered here. Here we have 4 attributes but they can be extended based on the complexity of the scenario.
The advantage of these formulae is that it allows for coverage of unknown entities as well as all the possible (including silent connections) in between them. In order to get the quality of the communication, we replace the values of each of the parameters and entity impact factors with the desired values of them and then divide them as:-

Quality = {(Actual System Connectivity) / (Desired System Connectivity)} * 100     (7)

A measure of atleast 50 – 75 % is considered satisfactory to be able to consider a scenario as having a high level of accuracy and learning capability. The Quality is a measurement of the understanding that was intended (and the ability to pursue further the tasks ahead including reduction of chances of confusion)

## IV. ASSUMPTIONS

1) The entities E will never have a value of 0 or 1 as entropy in an entity always exists and no entity can be perfectly stable or perfectly excited as per the laws of entropy.

2) There will always be obstacles or blocks to the success of the interaction

3) The ability to avoid these obstacles will decide the success of the interaction.

## V. METHODS

We collected information for all the above listed learning algorithms and then performed analysis on the same. We executed dry runs of software implementations of the above mentioned algorithms and found that they tend to confirm that their need of the entities to connect with each other. We found that each of these algorithms has the need to connect to its entities for establishing the communication, thereby always looking for information from past experience and logic using feed-forward and feed-backward propagation. We also found that these algorithms are always looking to find the "path" that will be best suited for them based on their inherent logic mechanism. Also, we found that whenever we put in an obstacle in the way of the path of the algorithm, it either stops or looks for a better way out. The entities that were mentioned clearly act as obstacles too as they are stepping stones for reaching the destination. These obstacles, present beforehand in the scenario of the algorithm, have compromised them and accepted these as required markings in order to reach the desired destinations. If these entities are not present, then they find the better path to them or stop the execution there itself. The following are the algorithms [4, 8] that were studied and analyzed to ensure their dependence on the law of connectivity.

*A.* SUPERVISED LEARNING ALGORITHMS:-

1) AODE

2) Artificial neural network e.g. Backpropagation

3) Bayesian statistics e.g. Naive Bayes classifier, Bayesian network, Bayesian Knowledge base.

4) Case-based reasoning

5) Decision trees

6) Inductive logic programming

7) Gaussian process regression

8) Group method of data handling (GMDH)

9) Learning automata

10) Minimum message length (decision trees, decision graphs, etc.)

11) Lazy learning

12) Instance-based learning

13) Nearest Neighbor Algorithm

14) Probably approximately correct learning (PAC) learning

15) Ripple down rules, a knowledge acquisition methodology

16) Symbolic machine learning algorithms

17) Subsymbolic machine learning algorithms

*B.* UNSUPERVISED LEARNING ALGORITHMS:-

1) Artificial neural network

2) Data clustering

3) Expectation-maximization algorithm

4) Self-organizing map

5) Radial basis function network

6) Generative topographic map

7) Information bottleneck method

8) Apriori algorithm

9) FP-growth algorithm

10) Single-linkage clustering

11) Conceptual clustering

12) K-means algorithm

13) Fuzzy clustering

14) Reinforcement learning e.g. temporal difference learning, Q-learning

15) Data Pre-processing

16) IBSEAD [22]

After forming the equations and the dry run flowcharts for the above algorithm implementations, we ran the example below to test their validity and the ability to deliver the desired results. Next, in each case, we followed the negative hypothesis to prove our theorem i.e. there is no need for a connection and any specific parameters for execution of tasks in machine learning. Thus, the above statement requires the removal of any connections from the scenario of any entity's interaction. In neural network, the example below was executed perfectly. However, we then ran the program by reducing one entity (per cycle of execution run) and blocked the connection. These connections were removed in two phases:-

1) The important connections were retained and the less important connections were removed first or blocked

2) The important connections were removed first or blocked.

This was done in two phases and the results were recorded. Interestingly, the results in the above scenarios reduced in quality with a drop in the connections available for interaction. Again, in the next step, we introduced a new connection, to replace the existing connections that were blocked. The results were analyzed. Interestingly, the quality of the results went higher with this introduction in most of the cases. In some cases, the connections that were blocked could not be replaced in impact and quality by the newly introduced connections. In the above scenarios, we have tried to compromise and ignore the disadvantages of each of the existing algorithms being studied and tried to find the best possible results that can be derived from them. We also manipulated the parameters mentioned in the above law for testing if there was any impact on the observations being recorded. The results were recorded and measurements made based on the equations. The results were then recorded and analysis done again to find if the law of connectivity in machine learning indeed holds true.

VI. THE CONCEPT OF CONFUSION IN MACHINE LEARNING

The law of connectivity explains the notion of confusion and self-connections. Confusion can happen when the self-connection of an entity (which could define the entity's self-understanding) is in conflict or not able to accept the connection with the destination entity. The effect of connection is in terms of the above hypothesis is given by

Max z = Connectivity (CB) where  (8)
z > 0, gives understanding and execution of steps and
z <= 0, which is a state of confusion

There are several reasons as to why confusion may happen in any given scenario (in the main entity's interaction):-
1) Missing information about a certain entity which is needed in the interaction. This may be a known, unknown or any hidden entity. Hidden entities have the highest chances of confusion in the scenario in which they are involved as they are not considered in the interaction by the main entity in the scenario.
2) Missing information about a certain path which is needed in the interaction. This may be an existing path or a silent connection (path)
3) Existing information within the self-connection (of the concerned entity) is having different values than the path of the concerned entity to the other entity.

For e.g. Consider a scenario having two entities A and B, with A being the main entity. The path AB is a connection between the main entity and the entity B. The main entity is having a self-connection path AA. We will consider this as confusion as path AB is not giving the same value as path AA. The main entity's understanding is different from what it is getting from the interaction with the Entity B. Thus based on the above equation, we get (assuming that the entities and the paths other than C (AA) will give positive effect in this example wrt the main entity A). For the entity A to be free from confusion and understandable in its interaction,

Connectivity (A) > 0 and Quality (A) > 50 %  (9)

Connectivity (A) = C (AB) - C (AA),  (10)

Where each connectivity measures from 1-10. If we assume that C (AB) = 4 and C (AA) = 5, we get:-

Connectivity (A) = -1.  (11)

Clearly, this means that there is some value of misunderstanding present in the above interaction. We can call this measurement as the quality of communication as this defines the level of understanding and the quality of meaning of the interaction which is done between the involved entities.

Quality(A)=Connectivity(A)/Desired Connectivity(A)  (12)

We take in this example Desired Connectivity (A) = 8 and the Quality is given by:-

Quality (A) = (-1 / 8) * 100                                              (13)

Ignoring the negative sign, we get

Quality (A) = (-) 12.5 %                                                  (14)

This means that the quality of communication is creating issues in the interaction and thus may affect the working of the scenario. This is clearly the case as there is a presence of confusion in the above example.

## VII. EXAMPLE

Consider the scenario where a few people are working in an office room. This room is divided into two teams with a half-wall in between. In the room, we consider each of these people as entities. Consider this scenario in which three people of Team A (the other being team B) are working with each other. As per the IBSEAD algorithm, there are hidden and unknown entities at work that can effect the interaction. Thus, these unknown entities can be noise and random people walking outside the room as they are beyond the control of the scenario and may have an effect on the scenario. The hidden entities are the entities that are known but may not be available for consideration directly i.e. the people from the Team B. In this scenario, we consider the situation where the persons Ea, Eb and Ec keep talking to each other such that:-

1) Ea is talking with Eb

2) Eb is talking with Ec

3) Ea and Ec are not talking with each other.

Now, Ea and Ec are known to be hostile and competitors of each other. We consider in this scenario a development at time 't' in which Ea is talking with Eb and Eb is talking to Ec. Here we find that Ec was trying to indirectly disturb Ea by distracting Eb in its interaction of the latter with Ea. As a result, the interaction between Ea and Eb suffered because of the noise that affected Ea. This noise had come from the interaction of Ec with Eb (Ec talking loudly with Eb and Eb hearing it). Thus, the noise has been considered as entity along with the possible effects of the unknown entities. Let us call the unknown entity as Eu (can be any external entity but we consider it as outside people here. In an ideal scenario, it can be difficult to find exactly who could be influencing the interaction under consideration here.) and the people from team B as Eh (we consider only one person form this team but can be more in an ideal situation. In such cases, multiply or add them as needed). In the end, Eb is started to get distracted and is trying to talk to himself (interacting with itself). There is some person outside the door because of which there was a big thud on the ceiling of the room (external unknown entity affecting problem here). Thus, we get the connectivity (for Eb being the host entity here)  as:-

Connectivity( Eb) = Q( Ec-Ea) + Q(Ea – Eb) + Q(Ec – Eb) + Q ( Eh – Eb ) + Q ( Eu – Eb) + Q( Eb-Eb)        (15)

Here the self silent connection of Q (Eb-Eb) is having negative effect. As per the equation (The result being on a scale of 1-10):-

$$C_i = F(E_i \Longleftrightarrow E_j) = \sum_{t=0}^{\infty} \sum_{i=0}^{n} \sum_{j=0}^{m} \sum_{k=0}^{p} \sum_{l=0}^{q} ((Imp) * E_{(i,k)} -- (Imp) * E_{(j,l)})$$  (16)

Consider the value of each of the entities to be Ci = 7. Please note that the entity Eb is talking to himself and thus has a silent connection [22] with him exists. Also, Q (Ec-Ea) is a silent connection and Q (Eb-Ec) is negative as. Thus, we get

C (Eb) = -7 + 7 + 7 + 7 + 7 – 7 – 7 = 7                         (17)

Now we find the ideal scenario in which all the entities are positively influence the scenario of Eb. That would give the value of the above equation as:-

C (Eb) = 7 * 8 = 56.                                               (18)

Thus, we now get the efficiency level of the communication as:-

Efficiency (Eb) = 100 * 7/56 = (1/8) * 100 = 13.5% (of what the ideal connectivity should be).                              (19)

This explains that the efficiency of the conversation is very low and that steps should be taken to make this positive. Lower reliance on silent connections will reduce the value of the same (though not eliminate the connection between them) or even making the entities become positive. For e.g. if Ea and Ec are friends then the negative influence would become positive and thus improve the connectivity. Such steps would allow higher efficiency and lower reliance on imagination and unwanted connections in the interaction. Thus the value of any connection Ei can range from {-1*(1 - 10)} to {+1*(1-10)}. A value of 50-75 % efficiency is needed to have a satisfactory level of connectivity and communication in the scenario. Note that the law of connectivity has played a significant role in helping us get the silent connections, self-connections and the unknown entities into consideration here. We use the above theorem here that there is always a connection between two entities with at least a self-connection being present. The earlier methods did not consider such an extensive coverage and also did not consider the state of the connections (positive or negative) for each of them. It is also known that imagination is a state in which the person believes that there is a connection between it and the desired entity object. When the actual entity meets the imaginary connection due to overlap, then the imaginary connection's value becomes positive. However, in the above example, it is negative as the overlap did not happen (what the entity imagined was not true, leading to negative value of the same).

## VIII. RESULTS

We executed the above mentioned algorithms and recorded the results as mentioned in the Method section. It was interesting to note that each of the algorithms required the urge or need to connect to the other entity. This is in line with the statement mentioned in the above law that entity requires the need to connect. Again, we also found that the unknown entity were already present in the scenario. However the entities were ignoring the presence of the same. The results kept changing based on the parameters that were covered. We also measured the results based on the parameters mentioned in the above law. It was found that these above mentioned parameters played an important role in affecting the execution of the above mentioned learning algorithms. This is because the blocking of important entities or paths reduced the quality of connectivity by 5%. In the next step, the introduction of new entities was done to measure if the law could stand for the explanation of novelty and surprise – some higher level human brain functions. We found that each algorithm was actually trying to find a way to reduce its dependency on the blocked path and thereby satisfactorily find the intended connectivity at the highest possible level of quality. Novelty is the ability to find new solutions to existing problems in the presence of no known or existing solutions. The above scenario, in which one connection is blocked and a new one introduced, also tried to explain this concept. Clearly, the above law of connectivity was satisfied by the above mentioned example and the equation tested. Using neural network algorithms and other existing methods would consider only the known and a few hidden entities in the above example. Thus, the lack of coverage would go down. The above mentioned attributes in the law of connectivity would be ignored leading to lower efficiency levels. Higher level functions like observation and imagination are dependent on these functions. Also, the positive/negative influence on the interaction is not considered well in neural relaxation models, which we have done so here. The result is that the final answer would go down in accuracy by 20% of the answer that we have achieved in the example here. The above example is actually very complex and that each of the given entity's silent connection would be considered. However, for the simplicity that is vital for using the formulae, we have considered only a subset of the same here. However, in reality, the law in this paper directs us to use all the entities to get the highest accuracy levels. Note that the silent connection also exists between the unknown entity Eu and Eb. Eb can only imagine that Eu is causing some problem for himself. Thus, this silent connection Q (Eb-Eu) can have negative influence with the connection being imaginary. However, when the entity gets the reliable information that Eu is actually causing the problem, then another connection between Eu and Eb is created. This is because the imaginary and actual connections come into existence and both will overlap significantly to give the desired accuracy levels. Thus, we would add this as a connection here in the above example also.

## IX. Advantages

There is a need for a fundamental law that will lay the base for all the learning algorithms in machine learning, which will act as the base for them. The law of connectivity clearly serves in doing this and assures that the future algorithms will follow the same.

1) This law clearly lay down the attributes that are needed to serve as a base for any type of learning

2) The law will ensure that all the learning algorithms are have a common mode of existence and help in better and more complex algorithms.

3) The higher levels of human brain activities such as observation, imagination and creativity are still to be fully implemented due to the absence of such fundamental laws in its processing. With the equation and the attributes clearly defined, it becomes easier for the computer to obtain the desired interaction.

4) The fundamental bases of the existing algorithms are clearly defined and serve to make them better.

5) This law helps in providing better efficiency and measurement of the entities under consideration.

## X. Advantages/need for the proposed system

There are many reasons why this law is needed and better in its application than the existing counterparts.

1) No such formulae exists till date that explains the higher levels function of the human brain e.g. confusion.

2) Neural Networks ignore the concepts of self-connections and repetitions in their scenarios and calculations – something which is a compulsory and routine feature in the real world interactions.

3) The presence of repetitive functions and other iterations is almost ignored by the existing algorithms. Moreover, the efficiency of these algorithms such as Neural Networks is reduced when such scenarios are considered

4) The features do not consider the presence of unknown entities – a concept that can help us explain the missing links about Novelty and concepts of Surprise and evolution.

5) Most of the attempts to map such high level functions have failed (e.g. the general problem solver) because they did not consider such fundamental concepts about the need to connect and the parameters involved in mapping the same.

6) Most of the algorithms in machine learning have failed to give a high execution rate i.e. they are known to work only in specific scenarios. This is in contrast to the ability of the human brain, on which most of the algorithms have been derived from. If the human brain is able to perform the above tasks well, then we can safely deduce that our understanding and implementation of the learning methods (which we call as learning algorithms) are far from the desired levels of human level intelligence.

## XI. Applications

1) This law is very useful in explaining complex scenarios where both hidden and unknown entities are involved.

2) This concept also helps in giving better insights into the secrets of imagination, creativity and other high level functions of the human brain, which still delude us.

3) This law can help in designing better algorithms that will take AI much higher in terms of artificial human brain functions for computers.

4) High level applications that failed till now, can be looked at now, with efficiency measurements, to improve the work further

5) This law should help in novel situations and in situations where no information about entities or entities from outside the environment is affecting the environment.

6) Neural Networks are based on the neuron structure of our brain. However, the brain is unable to handle multiple

tasks at the same time. We use this law to allow for such abilities in the machines as this law does not depend on neural networks or decision trees for its structure. The structure is actually very hybrid and needs polynomial level derivation of the formulae.

## XII. CONCLUSION

The law of connectivity clearly serves as the base for creating new as well as understanding existing learning algorithms in machine learning. The algorithms cannot exist without the law of connectivity being implemented (at least partially) by its desired entities. The desired entities will need to do that based on the obtained path and quality of communication. Thus, the law of connectivity holds true for machine learning algorithms and should thus be useful in embedding better and higher levels of intelligence in computers. In future, we would like to further improve this work to handle the abilities of scientific investigation and research by using this law and other complexities such as surprises better. We expect to further extend the above law for complex situations such as coincidence. We expect to implement this law of connectivity to implement higher levels of human brain abilities such as dreams, observation, etc better in computers. There are scenarios in which large groups with fuzzy knowledge and high levels of misunderstandings that can exist. There are also scenarios where the above equation may have entities which could act against the main entity. We aim to refine and implement a new learning algorithm based on the above obtained results and details of the law. We end by stating that the hypothesis of the law of connectivity in machine learning holds true. We hope that we these findings will help scientists further their work in this field and successfully implement the higher level human functions in machines for human benefit.

## ACKNOWLEDGEMENT

The author thanks his friends and family for this work. A special thank you to Prof. Uma Srinivasan for her inspiration and guidance.

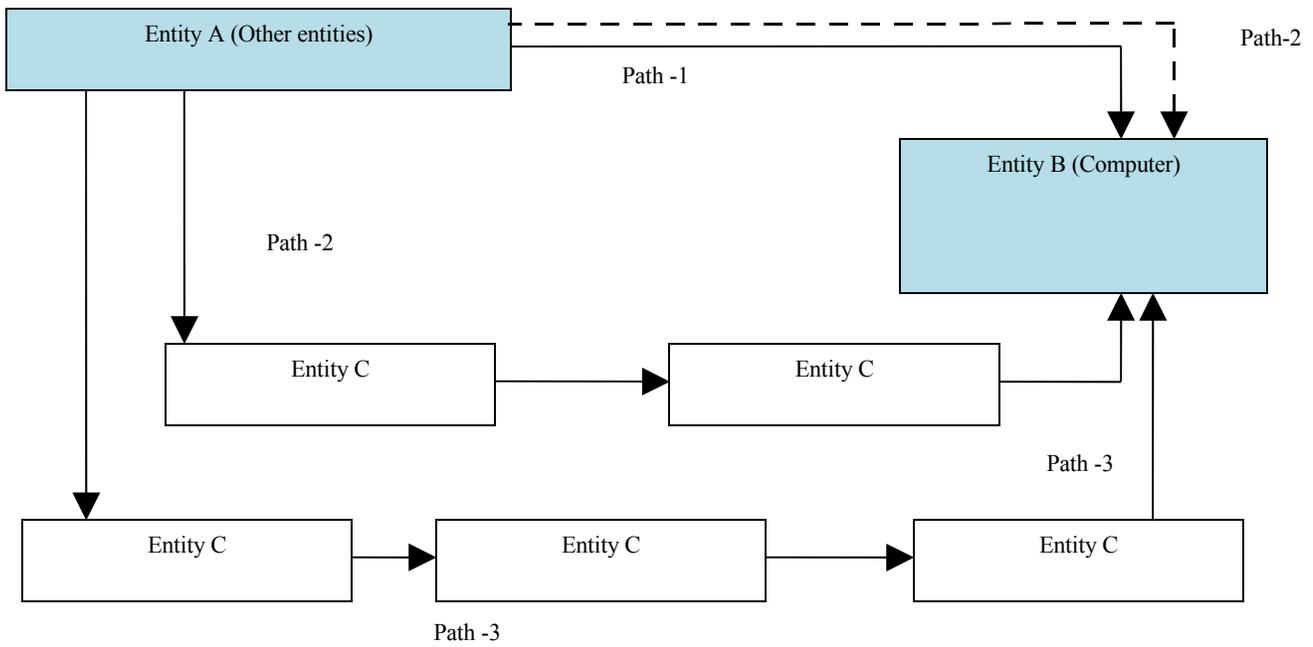

**Figure 1. Connectivity between entities**

The desired connectivity is between A and B, which are shown in blue. The other entities, which come in between the paths routed through them, are shown in white. Those in dashed lines are actually the silent entities while solid lines are the physically real connections.